# URBAN TRAFFIC DYNAMIC REROUTING FRAMEWORK: A DRL-BASED MODEL WITH FOG-CLOUD ARCHITECTURE

**Runjia Du**
Graduate Research Assistant, Center for Connected and Automated Transportation (CCAT), and Lyles School of Civil Engineering, Purdue University, West Lafayette, IN, 47907.
Email: du187@purdue.edu
ORCID #: 0000-0001-8403-4715

**Sikai Chen***
Visiting Assistant Professor, Center for Connected and Automated Transportation (CCAT), and Lyles School of Civil Engineering, Purdue University, West Lafayette, IN, 47907.
Email: chen1670@purdue.edu; and
Visiting Research Fellow, Robotics Institute, School of Computer Science, Carnegie Mellon University, Pittsburgh, PA, 15213.
Email: sikaichen@cmu.edu
ORCID #: 0000-0002-5931-5619
(Corresponding author)

**Jiqian Dong**
Graduate Research Assistant, Center for Connected and Automated Transportation (CCAT), and Lyles School of Civil Engineering, Purdue University, West Lafayette, IN, 47907.
Email: dong282@purdue.edu
ORCID #: 0000-0002-2924-5728

**Tiantian Chen**
Postdoctoral Research Fellow, Department of Industrial and System Engineering, The Hong Kong Polytechnic University, Hung Hom, Kowloon, Hong Kong
Email: tt-nicole.chen@connect.polyu.hk

**Xiaowen Fu**
Professor, Knowledge Management and Innovation Research Centre, and Department of Industrial and System Engineering, Hong Kong Polytechnic University, Hung Hom, Kowloon, Hong Kong
Email: xiaowen.fu@polyu.edu.hk

**Samuel Labi**
Professor, Center for Connected and Automated Transportation (CCAT), and Lyles School of Civil Engineering, Purdue University, West Lafayette, IN, 47907.
Email: labi@purdue.edu
ORCID #: 0000-0001-9830-2071

**ABSTRACT**

Past research and practice have demonstrated that dynamic rerouting framework is effective in mitigating urban traffic congestion and thereby improve urban travel efficiency. It has been suggested that dynamic rerouting could be facilitated using emerging technologies such as fog-computing which offer advantages of low-latency capabilities and information exchange between vehicles and roadway infrastructure. To address this question, this study proposes a two-stage model that combines GAQ (Graph Attention Network – Deep Q Learning) and EBkSP (Entropy Based k Shortest Path) using a fog-cloud architecture, to reroute vehicles in a dynamic urban environment and therefore to improve travel efficiency in terms of travel speed. First, GAQ analyzes the traffic conditions on each road and for each fog area, and then assigns a road index based on the information attention from both local and neighboring areas. Second, EBkSP assigns the route for each vehicle based on the vehicle priority (vehicle's proximity to intended destination) and route popularity (route's frequency of patronage). A case study experiment is carried out to investigate the efficacy of the proposed model. At the experiment's model training stage, different methods are used to establish the vehicle priorities, and their impact on the results is assessed. Also, the proposed model is tested under various scenarios with different ratios of rerouting and background (non-rerouting) vehicles. The results demonstrate that vehicle rerouting using the proposed model can help attain higher speed and reduces possibility of severe congestion. This result suggests that the proposed model can be deployed by urban transportation agencies for dynamic rerouting and ultimately, to reduce urban traffic congestion.

# INTRODUCTION

## Background

Urban traffic congestion continues to pose a persistent problem in several major cities due to increasing populations, vehicle ownership and travel demand. According to the 2020 annual report from the U.S. Department of Transportation, the total number of registered vehicles increased from 250 million to 273 million in 8 years (2010~2018) (*1*). The high levels of traffic congestion have adverse consequences on social wellbeing and economic productivity. Intelligent Transport Systems (ITS), which is largely rooted in information and communication technologies, provide various route-guidance methods that are particularly useful in dynamic traffic environments. Car-navigation systems including GoogleMap and TomTom use infrastructure-based traffic information to compute and prescribe traffic-cognizant shortest routes to their users (*2*). In addition, several large cities have deployed, on a wide scale, other traffic guidance systems including Variable Message Signs (VMS), to broadcast real-time traffic flow information. However, as the information is available to all drivers, these solutions may provide identical guidance for all vehicles with similar destinations simultaneously, and therefore tend to merely shift the traffic congestion to other locations of the road network, and the overall congestion issue remains unresolved (*3*). To address such "congestion-shifting" effects of congestion-mitigation initiatives, multi-route planning algorithms have been proposed in existing literature. One of these is simply to calculate K alternative routes and then randomly assign them to the vehicles (*4, 5*). However, such rerouting might yield a further inferior solution because the vehicles that are already close to their destinations may be randomly rerouted by the algorithm to take a longer detour. Therefore, in assigning the routes to the vehicles, it is vital to consider the priority of the vehicles. In addition, it is important to consider the "popularity" or, the frequency-of-use, of each route. As a general rule of thumb, vehicles with relatively lower priority should not be assigned routes that are assigned frequently (popular routes). Therefore, from a system efficiency perspective, it is more prudent for lower priority vehicles to be assigned routes with relatively lower popularity. Pan et al. used the Entropy Balanced k Shortest Path (EBkSP) algorithm to dynamically reroute vehicles and demonstrated that the algorithm can efficiently assign vehicles to appropriate routes thus addressing systemwide congestion without shifting congestion, with reasonably low computational effort (*5*).

## Fog-cloud architecture for multi-agent system

It is also worthy to recognize that urban transportation networks are extremely complex systems that are highly interconnected, and information exchange needs to be carried out in a very efficient manner. However, information exchange (communication) under a cloud-based computing environment in large networks can be time-consuming (high latency), because the information resources are located in the core of the network (*6*). Fortunately, fog information resources, unlike clouds (DataCenters), are located on the edge of the network, and by decreasing the distance from the core to the users, fogs can enhance communication efficiency.

Fog nodes (or, fog agents) refer to distributed fog computing entities that enable the deployment of fog services with processing and sensing capabilities (*7*). As shown in Figure 1, each fog node governs different regions (also referred to as "fog node areas") and collect regional data such as vehicle speed, vehicle location and vehicle density. With the multi-agent system (MAS) constituted by fog nodes, a dynamic rerouting framework can take advantage of the fog computing using the fog nodes to collect and exchange local information. In several literature, fog computing has been used to effectively assist in dynamic rerouting. Brennand et al., proposed an ITS architecture using fog nodes they termed "Fog RoutE VEhiculaR (FOREVER)" (*8*). In that application context, however, the lack of communication among the fog nodes could possibly lead the fogs to recommend routes that are only locally optimal. Also, Cao et al. designed a traffic congestion scheduling scheme using ITS architecture that incorporates fog computing. In their study, the fog nodes can communicate and share information to characterize overall traffic conditions, and the K alternative routes were calculated to prevent the same route from always being selected (*9*). In the Cao et al study, even though the fog nodes related to each other, the routes were still calculated locally. Moreover, compared with cloud, fog nodes have relatively weak computing capabilities. In yet another study, Rezaei et al. evoked a fog-cloud based architecture to guarantee that the vehicles are assigned the best routes globally using cloud computing to provide supplementary information where the local information from the fog node is insufficient (*10*). They demonstrated that a combination of fog and cloud can represent an efficient architecture that combines local information exchange and global route guidance.

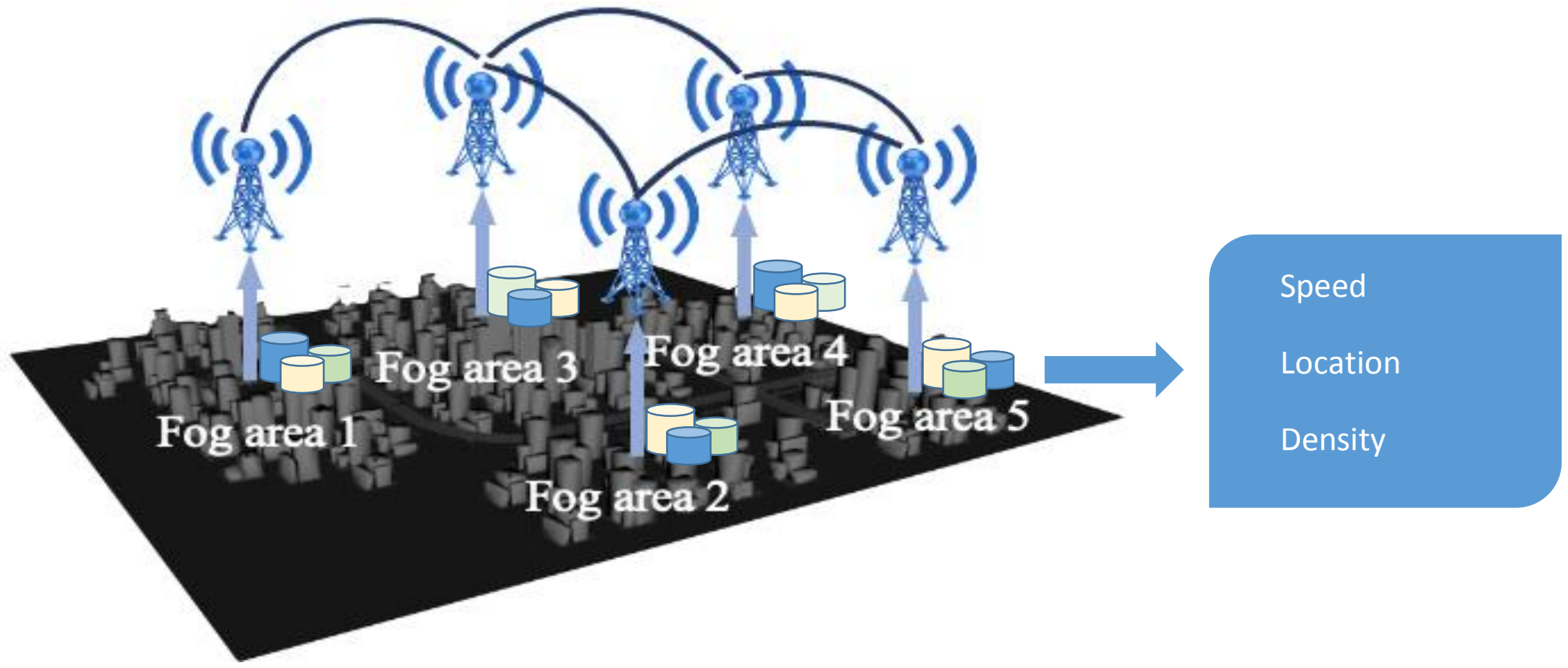

**Figure 1. Graph Structure of the Fog Nodes**

**Attention mechanism for deep reinforcement learning**

Evidence from the literature also suggest that, given the highly dynamic and complex nature of urban traffic systems, deep reinforcement learning (DRL) is a perfect solution for solving problems such as dynamic rerouting. Very few studies in the literature have used reinforcement learning (RL) to address the dynamic rerouting problem. Arkhlo et al. proposed a Multi-Agent Reinforcement Learning (MARL) to identify the best and shortest path between specified origin and destination nodes (*11*). Tang et al. generated an $A*$ trajectory rejection method based on multi-agent reinforcement learning (*3*). Yet still, there exist a few challenges in the use of DRL to address dynamic rerouting problems, particularly in a multi-agent environment: (i). multi-agent systems typically have large action space which have high costs of training time and convergence, and (ii). a large amount of information generated and collected in the network from different agents will affect the learning efficiency, because not all information is relevant. In this paper, MARL is used but combined with fog nodes. The fog nodes are regarded as agents which largely eliminate the action space. Fog agents' dependency and information flow can also be modeled using a graph. As shown in Figure 1, the nodes represent fogs, and the edges represent the connection between fogs. A deep learning-based fusion method with graph attention is proposed to generate the important features for each fog agent with due consideration of their neighbors.

Attention mechanism has been widely used to deal with variable sized inputs and focusing on the most relevant parts of the input to make decisions(*12*, *13*). Using a Graph Neural Network (GNN) combined with attention mechanisms, Veličković et al created an attention-based architecture to perform node classification of graph-structured data called Graph Attention Networks (GAT) (*14*).The hidden representations of each node in the graph are calculated by paying attention to the neighbors using a self-attention strategy. Since both local information and neighbor information are crucial for understanding the overall driving environment, a fusion method is needed to explicitly combine such information from different sources. Moreover, there is a need to differentiate the relative importance of input information based on the final decision. Thus, attention mechanism is essential for fogs to automatically "adjust the attention" to relevant information and GAT is an ideal candidate for this attention-fusion task due to its information fusion and attention ability.

There are a few research efforts that combined GNN and DRL. Jiang et al proposed a Graph Convolutional Reinforcement Learning (DGN) framework by using GNN as the encoder to learn representations between agents, then have the representations as input to a policy network (*15*).The joint trading of the encoder and policy network enabled the DGN agents to develop cooperative strategies. Chen et al built a Graphic Convolutional Q Learning (GCQ) framework by combining Graph Convolutional Neural Network (GCN) layer with Deep Q learning for Connected and Autonomous Vehicle (CAV) control. By generating the feature embedded mapping from GCN, and feed into Deep Q Network, the CAV can make lane change decisions in a sophisticated manner (*16*). Inspired by this recent research, this paper combines the GAT with Deep Q learning to help fog nodes generate decision on road index.

**Overview and organization of the paper**

The paper proposes a novel DRL-based dynamic rerouting framework based on a fog-cloud architecture. The fog nodes are the agents that collect and communicate the local information with the cloud. A deep learning-based fusion method with graphic attention is incorporated to generate important features for each fog agent considering information from both local and neighboring areas. A centralized multi-agent framework is then designed in the cloud which computes road index of different fog node areas using fog agents' fused features. After generating the road index, EBkSP method is used to assign appropriate routes to vehicles based on the computed road index while considering vehicles' priority and routes' popularity. The remaining parts of this paper is organized as follows: Section II presents the problem settings, which include the framework structure, DRL-stage settings (state space, action space, reward function) and routes assigned stage settings. The proposed methodology is introduced in greater detail in the Section III. Section IV, which is the experiment section presents the network settings, scenario features, and baseline model. Lastly, Section V summarizes the research, offers some concluding remarks, and suggests directions for future work.

## PROBLEM SETTINGS

The dynamic rerouting framework consists of two main stages. At the first stage (DRL stage), the network with fog paradigm is modeled as a graph whose nodes represent different fog nodes. The state, action and reward are defined to model the Markov Decision Process in the DRL stage for the agents to make decisions. By applying GAQ, road indexes for different regions are generated as the control variable for rerouting. At second stage (route assignment stage), vehicles calculate K alternative shortest paths based on the road index from the previous stage. Based on vehicle priorities and route popularities, the Entropy balanced method is applied to assign the appropriate route to each vehicle. Two types of vehicles are considered in this research (Figure 2): (a) Rerouting Vehicles (RV) (colored green) which are rerouted, and (b) Background Vehicles (BV) (colored white) which are not rerouted but are incorporated to add randomness and dynamics in the network. Both vehicles can be detected by fog nodes, and they have distinct origins and destinations.

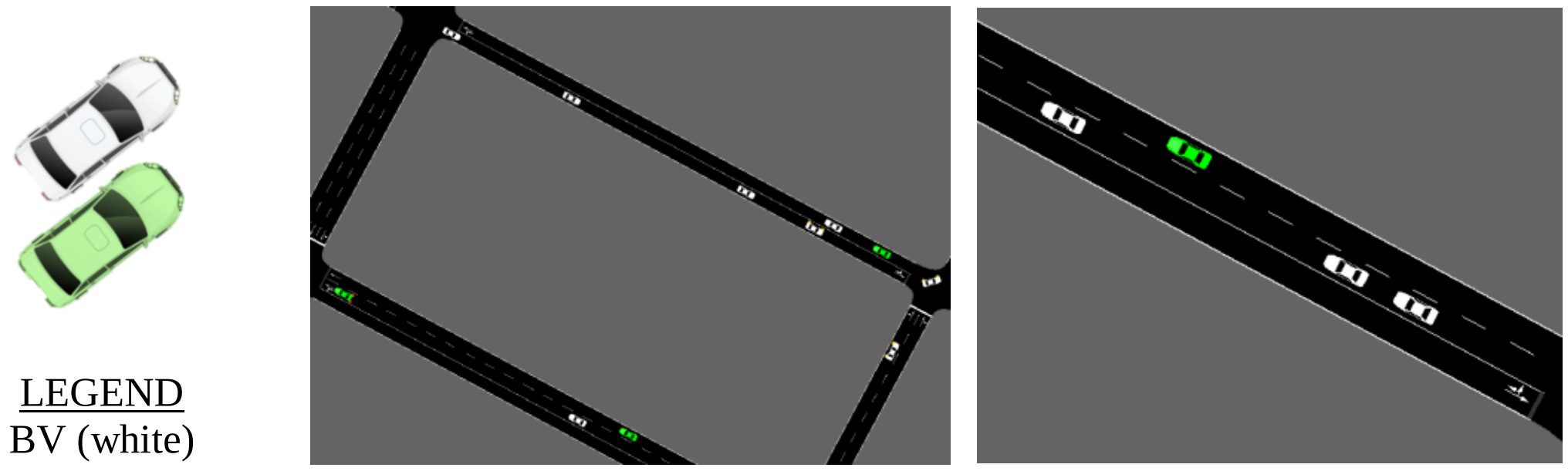

LEGEND
BV (white)
RV (green)

BVs and RVs in the urban network

**Figure 2. BV and RV in this Paper**

**The DRL stage**

In the DRL stage, four key factors are considered: Agent, State space, Action space and Reward function.
*Agent:* In this research, the fog nodes are considered as agents. At each timestep $t$, the agents choose actions $\{a_t^i\}$ based on existing policy and the current environment. Then, a reward is given to the agents based on their actions; this motivates the agents to strive for satisfactory results.
*State space*: The state space includes two parts: node feature at each time step $t$: $X_t$; and adjacency matrix at each time step $t$: $A_t$. At each $t$, RVs and BVs can be detected using fog nodes. Thus, network information is extracted from each fog area. Two types of information are included in the node feature matrix: average speed $\bar{v}_i$ and congestion condition $c_i$ (*17*).

- $\bar{v}_i = \frac{\sum_{k=1,\dots,n} v_k}{N_i}$ is the average vehicle speed of fog node area $i$, with $N_i$ equals to the number of vehicles in area $i$
- $c_i = \frac{\sum_{G=1,\dots,m} \frac{R^G_{num-veh} \times \tau}{R^G_{num-lane} \times R^G_{len}}}{M_i}$ is the average congestion level of the roads in fog node area $i$. $\{R^G_{num-veh}, R^G_{num-lane}, R^G_{len}\}$ represent the number of vehicles, length of road $G$ and number of lanes of road $G$, $M_i$ is the total number of roads in the area $i$. $\tau$ is a scalar to prevent $c_i$ from becoming too small.

The adjacency matrix $A_t$ is a binary matrix with dimension of $N \times N$, where $N$ is the number of nodes. $A_t$ reflects the information topology and dependency of the fog nodes. In this study, the graph of the road network is directed, but the graph of the fog layer (for information dissemination purpose) is undirected, and $A_{ij} = 1$ represents the existence of a connection between fog nodes $i$ and $j$.
*Action space*: for each time step, each fog node has a discrete action space representing the potential road index of the area: $a_i = \{0, 1, 2, 3, 4\}$. The action space for reinforcement learning is aggregated by all possible actions of each fog nodes: $\mathcal{A} = [a_i], i = 1, \dots, n$. After generating the road index, the actual weight $R^j_{weight}$ for road $j$ in area governed by fog node $i$ can be calculated based on the road index $\mathcal{R}^i_{index}$ and road vehicle density (number of vehicles):

$$\begin{bmatrix} R^1_{weight} \\ \vdots \\ R^j_{weight} \end{bmatrix}_i = \mathcal{R}^i_{index} \times \mathcal{T}_1 I + \mathcal{T}_2 \begin{bmatrix} R^1_{density} \\ \vdots \\ R^j_{density} \end{bmatrix}_i \tag{1}$$

Where $\mathcal{T}_1$ and $\mathcal{T}_2$ are balance terms to help avoid overwhelming of the road vehicle density on the road index or vice versa.
*Reward function*: In the reward function, we consider both reward and penalty based on average vehicle speed in the network. The purpose of the proposed dynamic rerouting framework is to maintain and enhance the RVs' efficiency. The speed change reflects a change in the traffic conditions. A drastic drop in the average vehicle speed is often symptomatic of congestion. Thus, the framework uses a speed increase reward and speed decrease penalty with threshold of 5 m/s (11 mph).

**Routes assigned stage**

After determining the actual road weight of the network, EBkSP is used to calculate routes for each RV based on their current location. All vehicle information is collected by the fog nodes and considered in the state space. Also, as mentioned above, the vehicle priority is important in choosing appropriate routes to avoid congestion shifts. Given a set of RVs: $RV = (RV_1, RV_2, \dots, RV_n)$ to be rerouted, the distance to the destination is used to compute the priority of the RVs. In this paper, we considered two different standards to calculate vehicles' priority:

- Priority1-Near: Based on the destination of RVs' current location to their destination, RVs that are nearer to their destinations are assigned hiher priority.
- Priority2-Far: Based on the destination of RVs' current location to their destination, RVs that are further to their destinations are assigned higher priority.

## METHODOLOGY

### DRL model architecture

In the DRL stage, the settings are multi-agent planning with centralized learning but decentralized execution (*16*). Each fog agent makes decisions at each timestep, and the target is to achieve a same given goal for all the agents, which improves the efficiency and avoid congestion of the rerouting vehicles in the network. The information attention is modeled with GAT and the decision processor used is Deep Q learning.

At each timestep $t$, vehicles (RVs and BVs) are detectable by fog nodes. The input of the model is the state $s_t$. The state is a tuple of $N \times F$ nodes feature matrix $X_t$ and $N \times N$ adjacency matrix $A_t$, $N$ is the number of nodes, and $F$ is the number of features in each node. There are two features considered in nodes feature matrix: (i) the average speed in the fog area $i$; (ii) the congestion conditions in the fog area $i$; fog nodes send their local information to the cloud and then the network node features are concatenated by fog nodes' information. During the information fusion process, the adjacency matrix is used to indicate the relationship between the fog nodes.

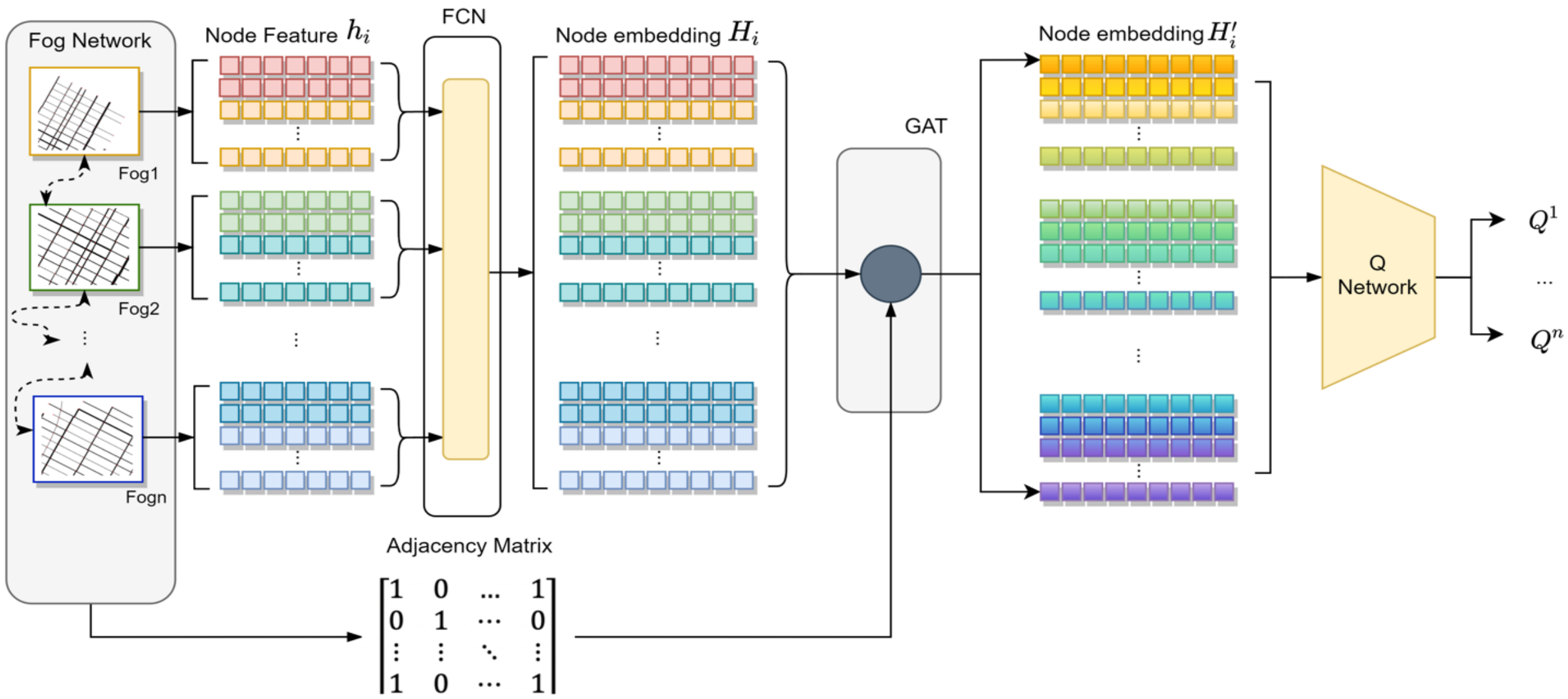

**Figure 3. DRL Model Architecture**

As shown in Figure 3, the model consists of the following parts: a fully connected network encoder, a GCN layer, the Q network, and the output layer. At each timestep $t$, the nodes feature matrix $X_t$ is used as the input to the FCN encoder $\varphi$ to generate node embeddings $H_t$ in $d$ dimensional embedding space

$$H_t = \varphi(X_t) \in \mathcal{H}, \mathcal{H} \subset R^{N \times d} \tag{2}$$

Then the graph convolution with attention mechanism is applied to the node embeddings $H_t$. Unlike the GCN layer, the GAT layer uses the attention mechanism to weight the adjacency matrix instead of using the normalized Laplacian.

$$H_t' = gat(H_t, A_t) = \alpha H_t W + b \tag{3}$$

$\alpha_{ij}$ is calculated using the attention mechanism and the adjacency matrix, it represents the coefficient for nodes $j \in \mathcal{N}_i$, where $\mathcal{N}_i$ represents a set of some first-order neighbors of node $i$ (including $i$). $T^{\top}$ is a weight factor that parameterize the attentional mechanism $T$:

$$\alpha_{ij} = \frac{exp\left(LeakyReLU\left(T^{\top}[(H_tW)_i \| (H_tW)_j]\right)\right)}{\sum_{k \in \mathcal{N}_i} \exp\left(LeakyReLU\left(T^{\top}[(H_tW)_i \| (H_tW)_k]\right)\right)} \tag{4}$$

The output of GAT layer is the node embedding $H_t'$, which is subsequently sent to a Q network $\rho$ to obtain Q values. Q values are used to evaluate the actions $a$. With $\hat{Q}$ representing the combined neural network blocks (FCN, GAT, and Q network), $\psi$ representing the combined weights, the model can be expressed as:

$$\hat{Q}_\psi(s_t, a_t) = \rho(H_t', a_t) \tag{5}$$

Experience Replay and Target Network (*18*) are used in the model training to enhance the learning efficiency. Also, the $\hat{Q}$ is trained on randomly sampled batches from replay buffer $R$ with size $B$ to obtain a stable performance. For each batch, the objective is to minimize the value of the loss function:

$$L_\psi = \frac{1}{B}\sum_t y_t - \hat{Q}_\psi(s_t, a_t) \tag{6}$$

Where $y_t = r_t + \gamma \max_a \hat{Q}_\psi(s_{t+1}, a)$. The architectures of different parts of the network are:

- FCN Encoder $\varphi$: Dense (32) + Dense (32)
- GAT layer $gat$: GATConv (32)
- Q network $\rho$: Dense (32) + Dense (32) + Dense (64) + Dense (64)
- Output layer: Dense (5)

Warm-up steps are added prior to the training to let the agents explore the environment thoroughly by taking random actions. After the warm-up steps, the training is performed by maximizing the reward and minimizing the losses. Algorithm 1 presents the detailed steps of this process.

**Algorithm 1. Graph Attention Q Learning**

**Initialize:**

Replay memory $R$

Joint weights $\psi$ and the target network $\hat{Q}_t = \hat{Q}_\psi$

**Warm up steps:**

For time step $t$ from 1 to $T_w$ **do**

- Random actions for each agent: $a_t = \begin{bmatrix} a^i \\ \vdots \\ a^n \end{bmatrix}$
- Gather and store $(s_t, a_t, r_t, s_{t+1})$ into the memory buffer $R$

**Training steps:**

For time step $t$ from $T_w + 1$ to $T$ **do**

- Update memory $R$ and choose new batch samples
- Take $s_t = X_t, A_t$ (Feature Matrix and Adjacency Matrix) and encode the node features into a node feature embedding $H_t = \varphi(X_t)$
- Apply graph attention mechanism $H_t' = gat(H_t, A_t)$
- Compute Q values for each action combination $a_t$: $\hat{Q}_\psi(s_t, a_t) = \rho(H_t', a_t)$
- Select optimal action $a_t^* = \operatorname{argmax}_{a_t} \hat{Q}_\psi(s_t, a_t)$
- Apply $a_t^*$ to the network and then obtain the reward $r_t$ and next state $s_{t+1}$
- Add the $(s_t, a_t^*, r_t, s_{t+1})$ into the memory buffer
- Move from state $s_t$ to state $s_{t+1}$
- From replay memory buffer $R$, get a random batch size $B$
- For each training examples with the batch, the target of Q value $y_t$ is calculated:
    - If $s_{t+1}$ is not done: $y_t = r_t + \gamma \max_{a_t} \hat{Q}_\psi(s_{t+1}, a_t)$
    - If $s_{t+1}$ is not done: $y_t = r_t$
- Losses are calculated by the loss function: $L_\psi = \frac{1}{B}\sum_t y_t - \hat{Q}_\psi(s_t, a_t)$
- The target network is updated based on the target updating frequency value

**Routes assigned model architecture**

In the vehicle route assignment stage, a local search method is applied to assign the proper routes to the RVs. Given the RVs set: $RV = \{RV_1, RV_2, \dots, RV_n\}$ At each time step $t$, the updated road weights for each road $\left\{R^j_{weight}\right\}_{j=1 \dots N_{total}}$ ($N_{total}$ is the total number of road segments in the network) are used to calculate K shortest alternative routes for each RV based on their currently-assigned road:

$$r_j \in \{kSP\}_i = ksp(R_{current}, RV_i) \tag{7}$$

$R_{current}$ represents the currently assigned road of $RV_i$. RVs' priority set $\mathcal{P}$ is obtained by the distance between their current location and the destination ($D_{RV_i}$) as shown in equation (8): Vehicles closer to their destination have higher priority, vice versa.

$$\mathcal{P} = (D_{RV_i}, RV_i) \tag{8}$$

Using $D_{RV_i}$, the priority of the RVs can be determined. The first $x$ vehicles are categorized as the "high priority" set: $RV_h$ ($x$ can be changed at the stage of model training), the rest of the vehicles are placed in the low priority set: $RV_l$. For the RVs with high priority, the shortest path in their $\{kSP\}$ set will be assigned to them. However, for the RVs with low priority, there is a need to calculate the popularity of their assigned routes, to ensure there will no congestion shift. The final assigned route is chosen from the K shortest alternative routes, and this prevents the vehicle with the final assigned route (the least popular route) from an excessively lengthy detour. The route popularity is defined as:

$$E(r_j) = -\sum_{i=1}^{n} \left(\frac{fc_j^i}{N_{r_j}}\right) \ln \left(\frac{fc_j^i}{N_{r_j}}\right) \tag{10}$$

Where $N_{r_j}$ is the number of roads in the route $r_j$. $fc_j^i$ ($i = 1, \dots, n$) is the road-weighted footprint of road $i$ in route $j$, which is calculated from: $fc_i = n_i \times \omega_i$. $n_i$ represents the total number of vehicles assigned to the routes that include road $i$, $\omega_i$ is a weight associated with road $i$ considers length, lane numbers and average free flow speed: $\omega_i = \left(\frac{len_{avg}}{len_i}\right) \times lane_i \times \left(\frac{V_{favg}}{V_{fi}}\right)$.

Algorithm 2 presents the detailed route assignment algorithm.

---

**Algorithm 2. Route assigned by EBkSP**

---

**Get Roads Weights:**

For $RV_i$ in set $RV$ **do**

- Find K-alternative shortest path based on current road: $\{kSP\}_i = ksp(R_{current}, RV_i)$
- Calculate the priority based on current location and add to the set $\mathcal{P} = (D_{RV_i}, RV_i)$

---

**Sorted Priority:**

Based on $sorted(\mathcal{P})$, let the top priority RVs into set $RV_h$ and others are low priority RVs into set $RV_l$

---

**Route Popularity:**

For $RV_m$ in set $RV_h$ **do**

- Assign the shortest route: $r_m^* = min\{kSP\}_m$ to $RV_m$
- Update the road weight footprint: $fc_i$

For $RV_n$ in set $RV_l$ **do**

- Based on the updated footprint, for the routes in $\{kSP\}_n$, calculate: $Pop(r_j) = e^{E(r_j)}$
- Assign the least popular route: $r_n^* = min\{Pop(r_j)\}$ to $RV_n$
- Update the road weight footprint: $fc_i$

---

## EXPERIMENT SETTINGS

The proposed framework is implemented in a simulation environment using SUMO (Simulation of Urban Mobility), which is an open-source simulator with well-defined vehicle parameters and vehicle controller (*19*). The training network is the Manhattan network (Figure 4 (a)) that is imported from OSM (Open Street Map) (*20*) and cleaned in SUMO, which is deployed in a multi-agent environment. As shown in Figure 4 (b), there are six fog nodes covers different regions of the network. To have the equivalent information collection of different fog node areas, each fog node covers about 50 roads and the roads in the network share. Simulator parameters, training parameters and baseline models used in the experiment are discussed in detail in a later section of this paper.

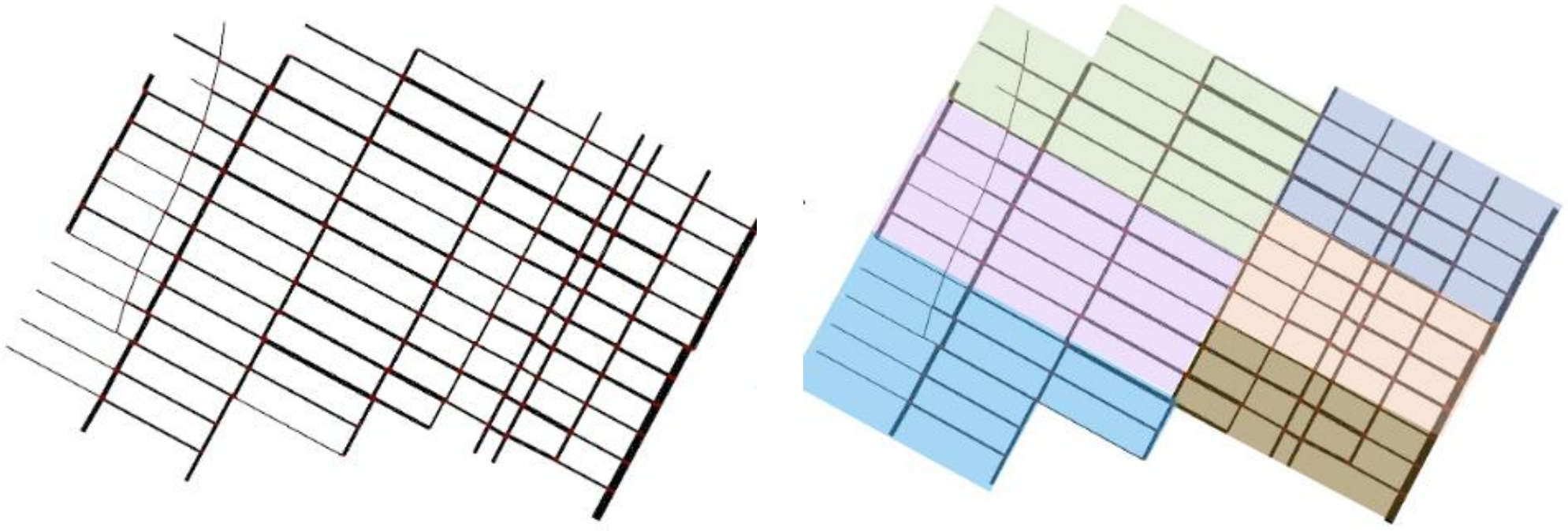

(a) Base network (b) Regions assigned to the fog nodes

**Figure 4. Case Study (Manhattan) Network With Fog Nodes**

### Simulator parameters

In SUMO, the driving simulation environment, network features, simulation scenario parameters, vehicle control parameters need to be defined well, based on the research problem:

*Network features*

A $5.926km^2$ area that extracted from the Manhattan area is used in this research, the network includes 287 edges (roads) and 120 nodes (junctions). The network structure is the same as that of the real world. There are multiple road types in the network: 2-lane roads, 3-lane roads, 6-lane roads, and 7-lane roads. Both one-way and two-way roads are included. The speed limit is reflective of the actual real-world conditions as evidenced by data from an open street map. The speed limit varies due to the different road types and ranges from 11 m/s to 28 m/s (average 15 m/s (30 mph)).

*Scenario parameters*

To increase the complexity and to mimic the dynamic nature of the urban road network enviroment, BVs (colored white) enter the study area from multiple areas with different travel patterns and destinations: (a) from right to the left, (b) from left to the right, (c) from the middle to the top, and (d) from the middle to the bottom; RVs (colored green) enter the map from 3 roads located on the right of the network (two from the top, one from the bottom) and two different destinations are located on the left of the network (one from the middle, the other from the bottom). At the training stage, the inflow rates of the BVs and RVs are both specified as 100 veh/hr. At the testing stage, the inflow rates are changed according to the number of BVs and RVs. Mixed traffic with both rerouting (RVs) and non-rerouting vehicles (BVs) needs to be focused on, which is more realistic in the real world (*21–25*). The total numbers of RVs and BVs in the training stage are 1000 with a 0.1 rerouting ratio, which is calculated by $\frac{RV}{BV}$, while different rerouting ratios will be tested in the testing stage.

*Vehicle control parameters*
The vehicle control in this paper includes vehicle behavior control and routing control. The vehicle behavior control includes car-following control and lane-changing control. In this study, both BVs and RVs use SUMO's built-in car-following and lane-changing controllers. However, the routing controller for RVs is based on proposed GAQ-EBkSP model, while the routing controller for BVs is simply EBkSP.

*Vehicle priority parameters*
As mentioned in a previous section of this paper, we consider two different ways to calculate vehicle priority (priority1-near and priority2-far). At the training stage, both priority1-near and priority2-far are trained. Moreover, as mentioned in "*Routes Assigned Model Architecture*" section of this paper, the length of the high priority set can be changed. Thus, at the training stage, different lengths of the high priority set will be implemented.

*Training parameters*
In the model training section, approximately 800 epochs are trained, with the first 200 epochs as warm-up stage. When training starts, transition batches of size 32 are sampled and put into the model. The optimization parameters used in this research is Adam (*26*) which has initial learning rate $\gamma = 10^{-4}$.

*Baseline models*
In this research, the baseline model is the EBkSP rule-based model that the RVs will be rerouted without the training stage. The road index will not be considered to taking any effect on the rerouting, the road weight will only care about the density of the vehicles. The rule-based model is implemented to compare with the proposed GAQ-EBkSP framework.

## RESULTS

### Training stage
At the training stage, the two priority standards (priority1-near and priority2-far) are specified under the proposed GAQ-EBkSP framework. As shown in the reward curves Figure 5 (a), priority1-near overperforms priority2-far. Based on the average reward lines that are calculated after convergence (after 400 episodes) of both priority standards, the average reward of the priority1-near scenario is higher than that of priority2-far by approximately 300 units. In this paper, the maximum number of steps for one episode is set to be 10. Typically, 10 steps per episode is adequate for all the RVs in the network to complete their trips. Therefore, if the maximum number of steps is reached in one episode, it means that rerouting vehicles are unable to complete their trips within the specified maximum number of steps. One evident reason could be that some of the RVs encounter severe congestion with BVs. Figure 5 (b) presents the probability of reaching the maximum episode steps of different priority standards throughout the training period. Clearly, as the training progresses, both priority standards show some progress, the probability of reaching the maximum episode step is lower. This means that the GAQ-EBkSP framework effectively prevents the rerouting vehicle from encountering severe congestion. On the other hand, using priority1-near scenario, there is 13% lower probability (compared to priority2-far) to encounter severe congestion.

With the exception of different priority standards, the length of the high priority set, which determines the number of high priority RVs is also crucial for the training stage. Therefore, we implemented three different set lengths for the high priority case: {5,10,15} under the priority1-near standard. As shown in Figure 5 (c), on the basis of the average reward lines (calculated after convergence), the performance of high priority length 5 (average reward of 893) is much worse compared to that of length 10 (which indicated an average reward of 1420) and that of length 15 (average reward of 1371). Even though the average rewards of high priority set length 10 and 15 are close, 10 is still a superior choice. This is because when the high priority set length is relatively large, almost all the vehicles in the network choose the shortest route without considering route popularity, which leads to congestion shifts from one part of the road network to another. As shown in Figure 5 (d), the probability of getting severe congestion using 15 as the high priority set length is higher than that associated with a set length of 10. Throughout the training process, the combination of priority1-near as the priority standard and 10 as the high priority set length overperforms other combinations. Therefore, this combination is used in the proposed model.

Figure 6 presents the episode reward curve of EBkSP (baseline mode) and GAQ-EBkSP (proposed RL model). The rule-based model without training shows a fixed value, that is, there can be no improvement in the reward. Also, the average reward of the proposed model is approximately 700 inuts higher than the baseline model.

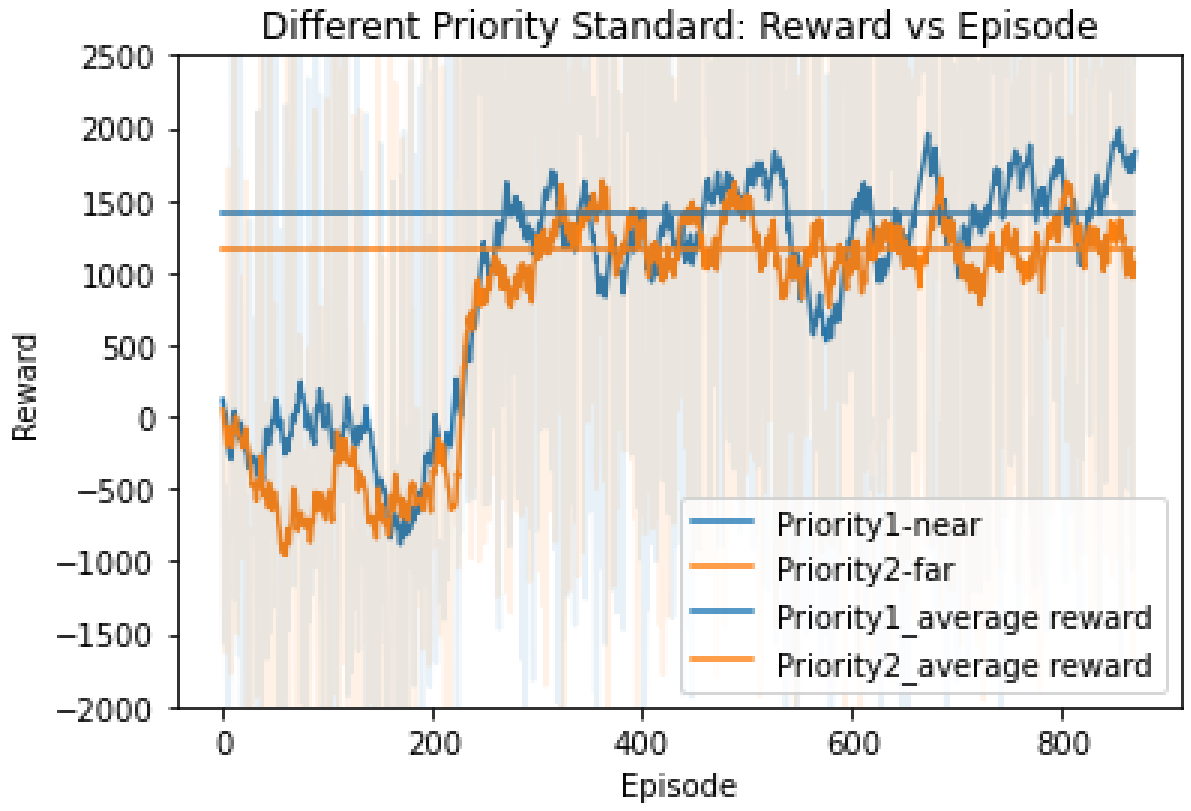

(a) Episode reward curve of Priority1-near and Priority2-far

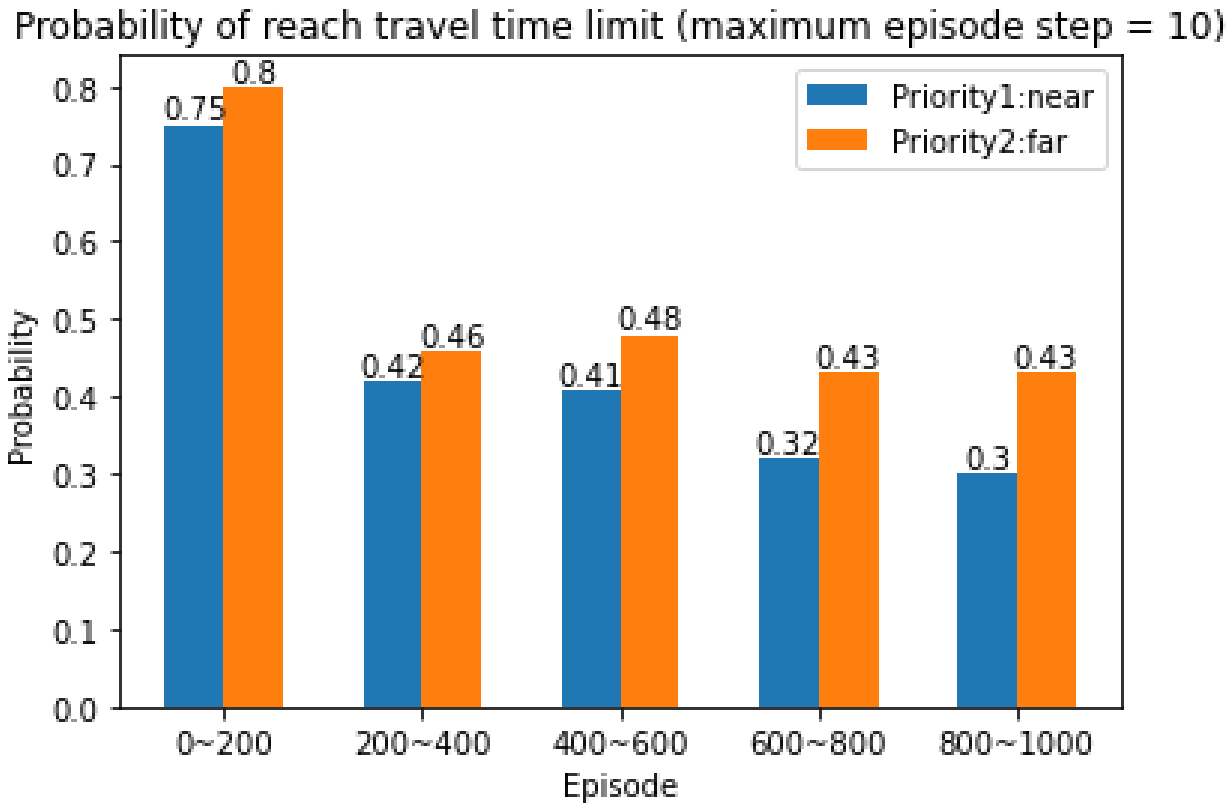

(b) Probability of reach travel time limit of Priority1-near and Priority2-far

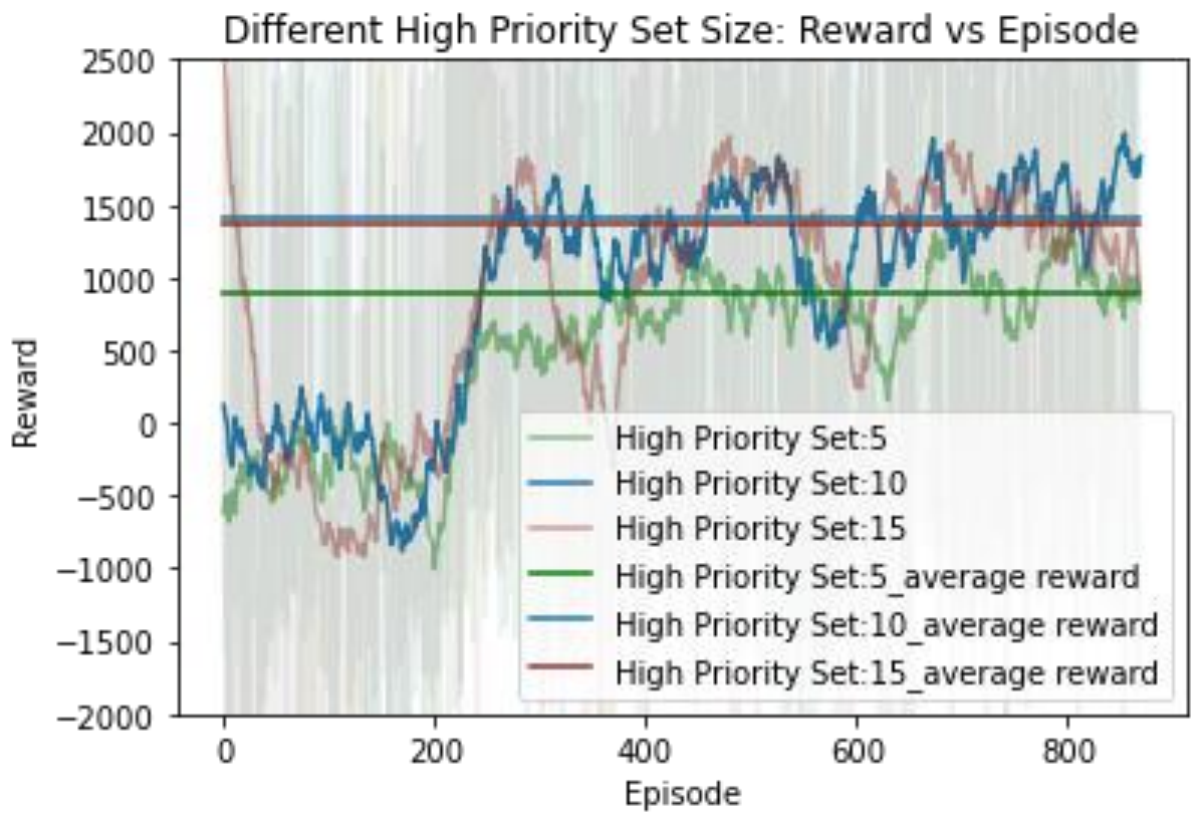

(c) Episode reward curve of Different high priority set size

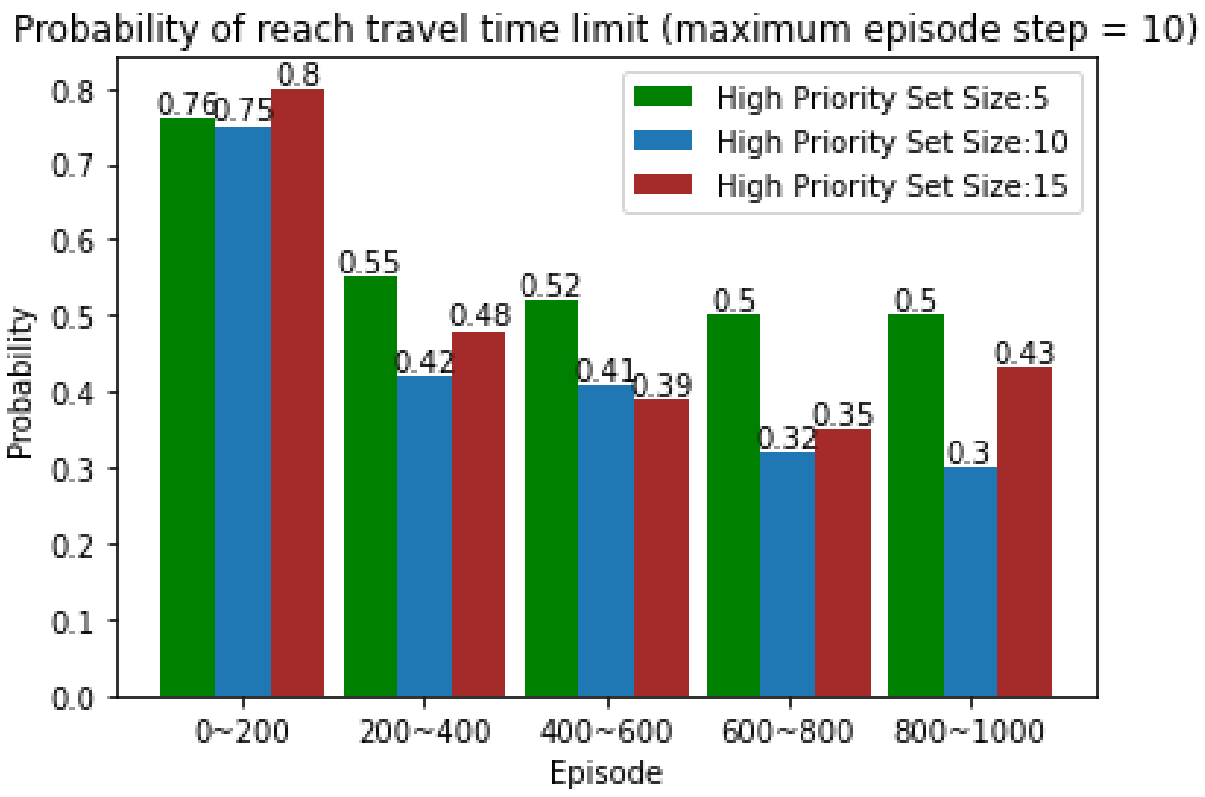

(d) Probability of reach travel time limit of Different high priority set size

**Figure 5. Training Performance of Different Priority Standards**

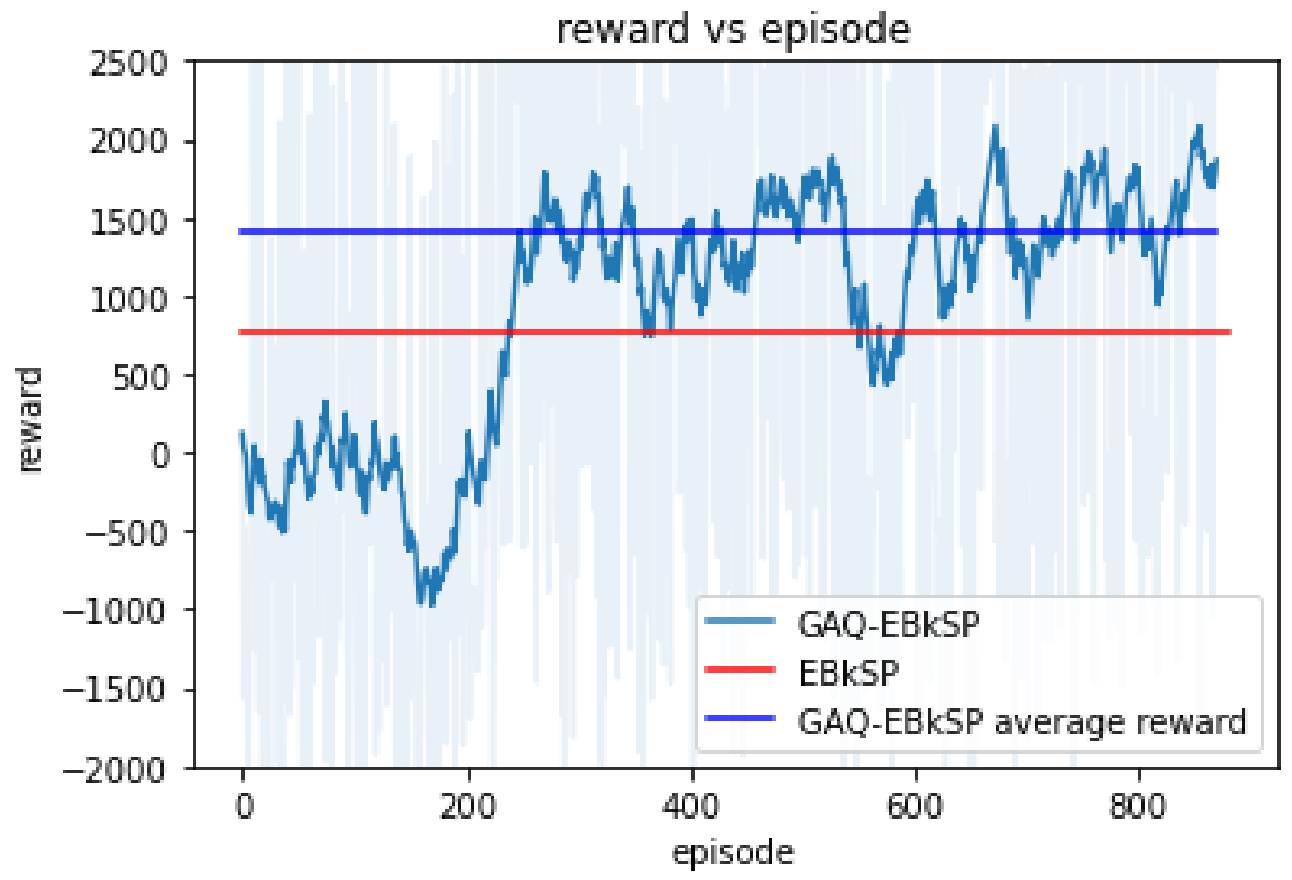

**Figure 6. Reward Comparison for Each Episode (Proposed vs. Baseline)**

**Testing stage**

Two important factors were considered in the testing stage: (i). the rerouting ratio (ii). total number of vehicles in the network (BV + RV). For the rerouting ratio test, the total number of the vehicles was 1000. Five different scenarios different ratios range from 0.1~0.9 were tested: 0.1 (900 BV and 100 RV), 0.3 (700 BV and 300 RV), 0.5 (500 BV and 500 RV), 0.7 (300 BV and 700 RV), 0.9 (100 BV and 900 RV). To increase the number of RVs in the network per unit time, the inflow parameter: vehicles per hour is increased. The number of RVs is also increased. To implement the total number of vehicles test, we set the rerouting ratio as a fixed number $r_{RV}$ and change the vehicles' number. Three scenarios with different total number of vehicles are tested: 1000 ($1000(1-r_{RV})$ BV and $1000r_{RV}$ RV), 1500 ($1500(1-r_{RV})$ BV and $1500r_{RV}$ RV), 2000 ($2000(1-r_{RV})$ BV and $2000r_{RV}$ RV). The performance metrics which are the average speed and the probability to encounter severe congestion, reflect the efficiency of the proposed method under the different scenarios.

As shown in Figure 7 (a) and Figure 7 (b), the proposed GAQ-EBkSP model outperforms the baseline EBkSP model under different ratios in both average speed and probability of reaching travel time limit (probability of the RVs encountering severe congestion). The average speed is an important metric that reflects the efficiency of the dynamic rerouting framework. In Figure 7 (a), the baseline model reroutes the vehicles based on the current density of the network while there is no control only input to tune the density between different road sections. As a result, the vehicles have no "future planning" and cannot choose superior routes jointly. This is the main reason why the proposed model achieves a higher reward. Particularly, this is observed where the ratio of rerouting vehicles is lower (a low ratio means that there are more background vehicles that are not under rerouting control), and in such cases, it is more likely to encounter severe congestion. As shown in Figure 7 (b), when the rerouting ratios are relatively low (≤30%), the average probability of encountering severe congestion is 0.35 when the proposed model is used, while the baseline model has an average probability of 0.55 which is 20% higher than that of the proposed model. When the rerouting ratios increases, the probabilities of encountering severe congestion are lower when either model is used. Yet still, even in the case, the proposed model outperforms the baseline model.

Based on Figure 7 (a) and Figure 7 (b), the worst case is when the rerouting ratio is 0.3. Thus, we use 0.3 as the fixed rerouting ratio in the test that investigates the effect of a different total number of vehicles. As shown in Figure (c) and Figure (d), the proposed model has superior performance with regard to both the average speed and probability of RVs encountering severe congestion. As Figure 7 (c) indicates, the proposed model exhibits a higher level of robustness compared to the baseline model. When the total number of vehicles is 1500, the average speed of the RVs under the proposed model is still high, indeed, higher than the scenatio with 1000 vehicles. This because as the number of RVs increases, some of the fast-moving RVs overtake the slow-moving RVs. In the Figure 7 (d), the probability of achieving the travel time limit is 0.4, which is the same with the scenario that has 1000 vehicles. However, the average speeds of RVs in the baseline model in first two scenrios are both lower than 10m/s. When the total number of vehicles reaches 2000, the average speed of RVs in the proposed model is about 7m/s higher than that of the RVs in the baseline model. Moreover, when using the baseline model, the probability of RVs encountering severe congestion is 100%, while this probability is only 65% when the proposed model is used.

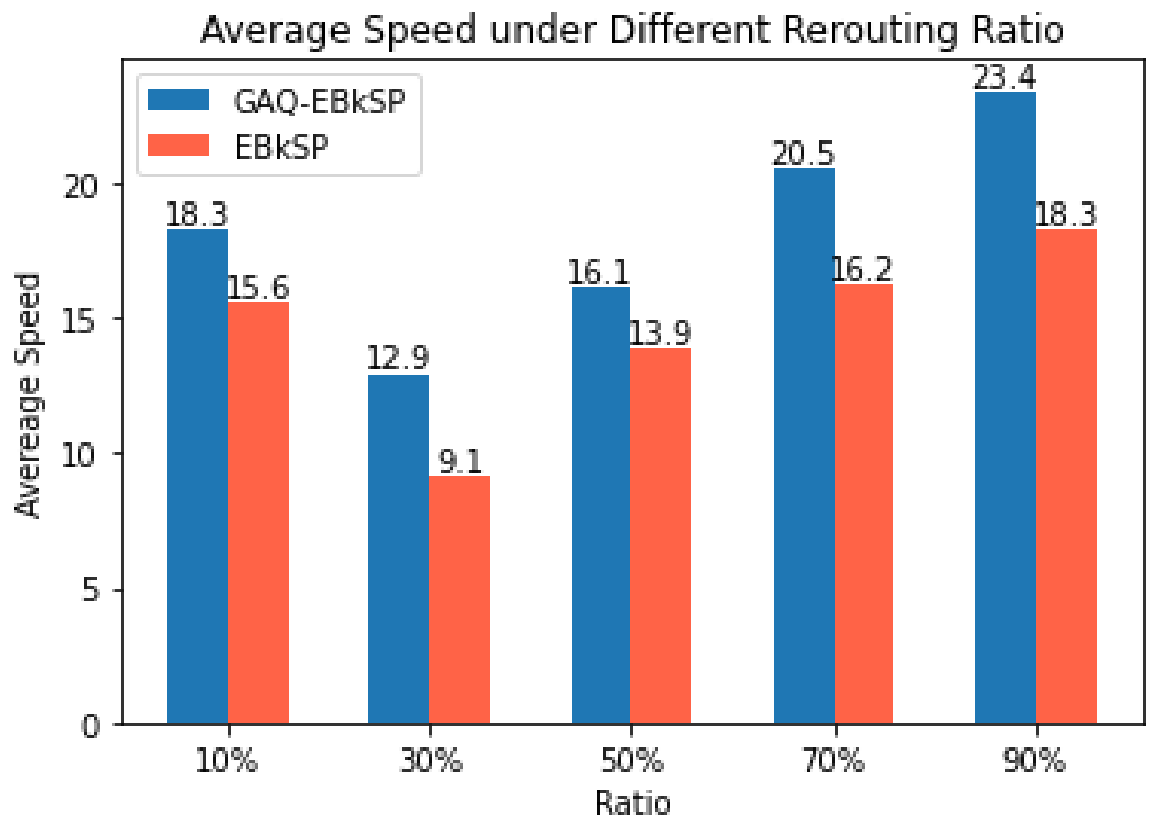

(a)Average speed for the Rule-based and RL-based models under different rerouting ratio

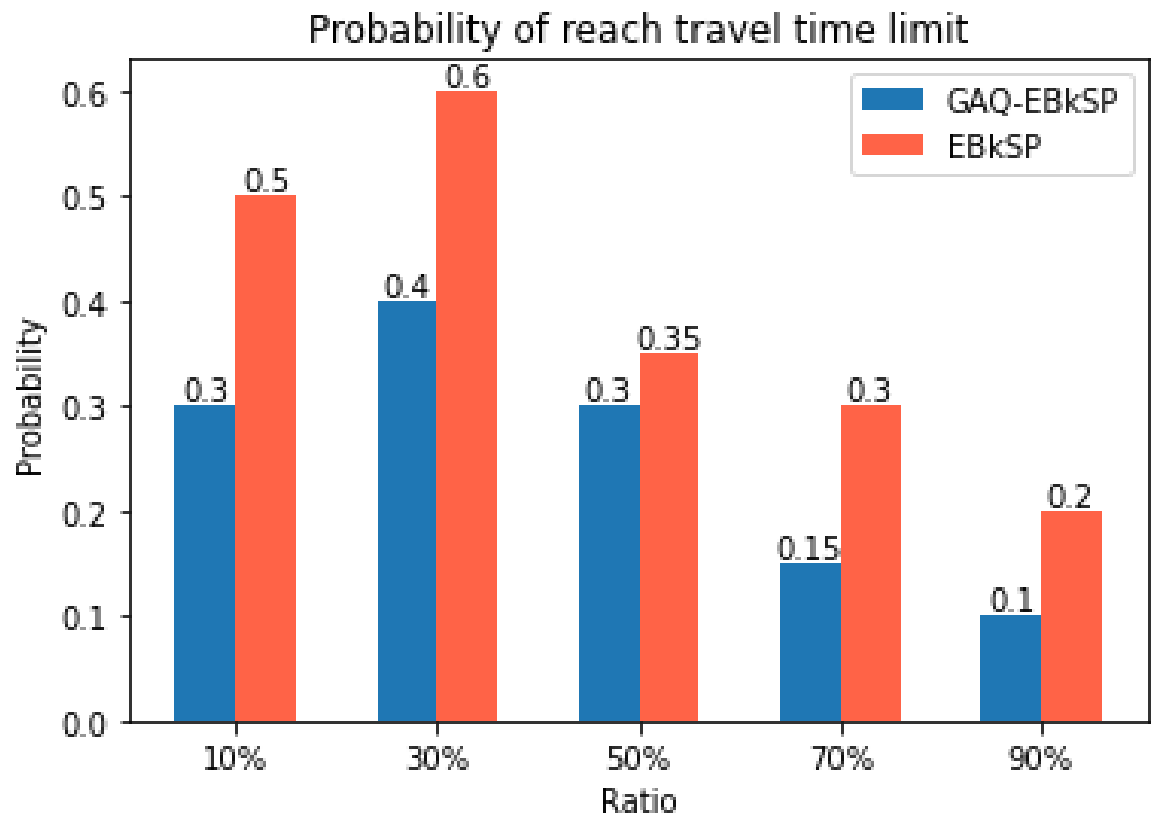

(b)Probability of reach travel time limit under different rerouting ratio

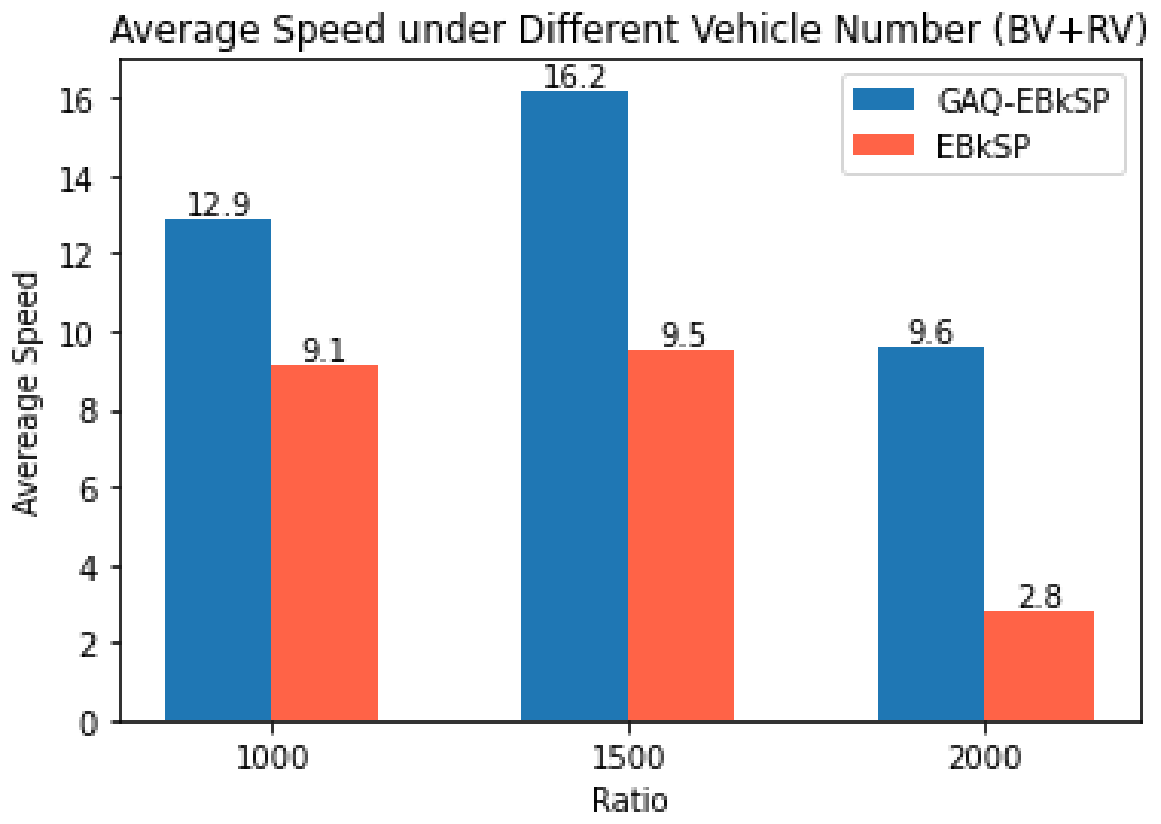

(c)Average speed for the Rule-based and RL-based models under different total number of vehicles

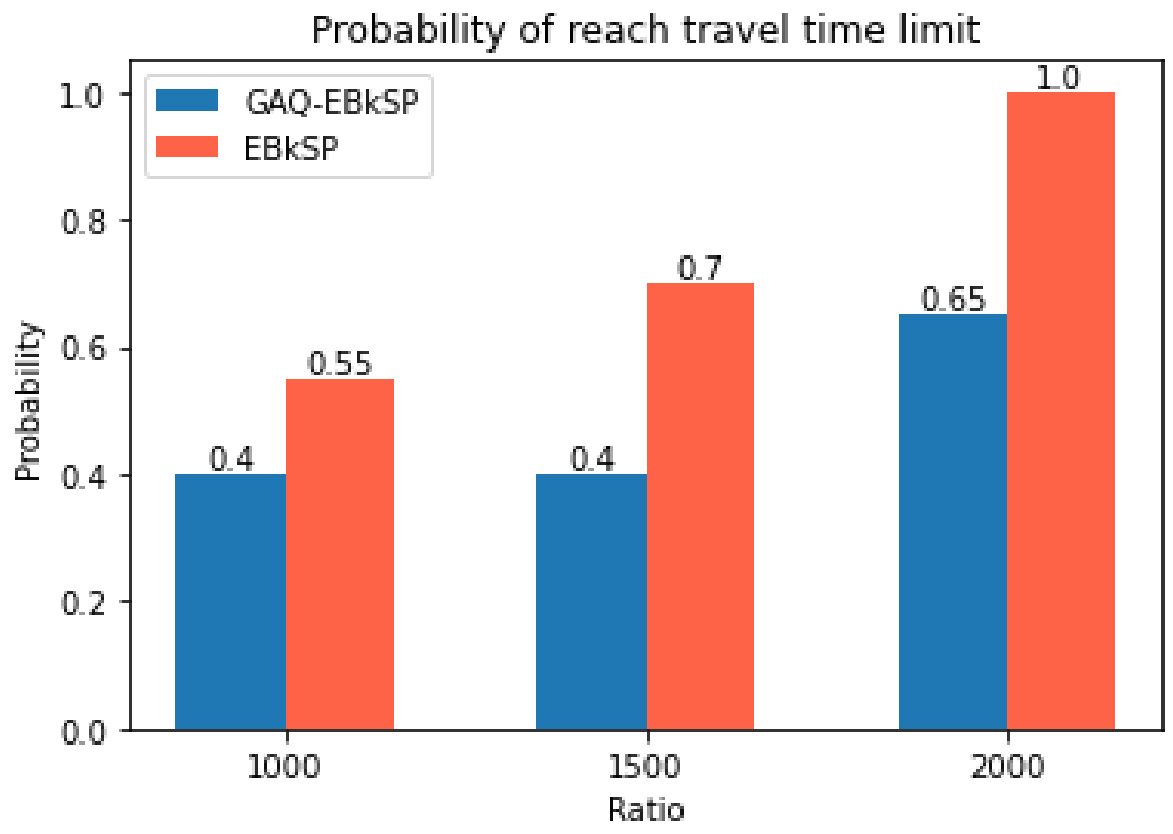

(d)Probability of reach travel time limit under different total number of vehicles

**Figure 7. Testing Performance of Rule-Based and Rl-Based Models Under Different Scenarios**

## CONCLUSION

In this paper, a DRL (GAQ)-EBkSP model based-on fog-cloud architecture is proposed to dynamically reroute the vehicles in large transportation networks. The fog nodes are the agents that learn the road weight of different regions, and the EBkSP method is used in cloud layer to search the proper routes for the vehicles based on the vehicle priority and route popularity levels. Moreover, the large action space problem in multi-agent system is solved by regarding fog nodes as agents instead of roads (each fog node covers an area of regions with multiple roads, which shrink the agents' number a lot). Furthermore, the paper applied a graph attention mechanism to fuse information and extract relevant information to enlarge the learning efficiency. The cloud layer helped ensure that the assigned routes are not only locally optimal, and that the routes are assigned to the vehicles based on their priority and routes' popularity. A region in mid-Manhattan, New York, is used as the experiment network study area, and different levels of the rerouting ratios (0.1~0.9) and the total number of vehicles (1000, 1500, 2000), are tested. The testing results suggest that the proposed model (GAQ-EBkSP) outperform the baseline model (EBkSP) in terms of average speed and probability of reaching the travel time limit in various scenarios.

The fog nodes layer plays a crucial rule in the proposed framework. In this research, six fog nodes are used to cover the network. However, different numbers of fog nodes and different fog node sizes are expected to influence the rerouting decision. Thus, in the future work, the impacts of different number of fog nodes and different fog node sizes can be studied. Moreover, similar RL-based model such as GCQ and LSTM-Q, can be used as baseline models at the testing stage to compare the results.

## ACKNOWLEDGMENTS

This work was supported by Purdue University's Center for Connected and Automated Transportation (CCAT), a part of the larger CCAT consortium, a USDOT Region 5 University Transportation Center funded by the U.S. Department of Transportation, Award #69A3551747105. The map data is from OpenStreetMap contributors and is available from https://www.openstreetmap.org. The contents of this paper reflect the views of the authors, who are responsible for the facts and the accuracy of the data presented herein, and do not necessarily reflect the official views or policies of the sponsoring organization. This manuscript is herein submitted for PRESENTATION ONLY at the 2022 Annual Meeting of the Transportation Research Board.

## AUTHOR CONTRIBUTIONS

The authors confirm contribution to the paper as follows: all authors contributed to all sections. All authors reviewed the results and approved the final version of the manuscript.